# Competitive Coevolution through Evolutionary Complexification

**Kenneth O. Stanley**                                    KSTANLEY@CS.UTEXAS.EDU
**Risto Miikkulainen**                                       RISTO@CS.UTEXAS.EDU
*Department of Computer Sciences*
*The University of Texas at Austin*
*Austin, TX 78712  USA*

## Abstract

Two major goals in machine learning are the discovery and improvement of solutions to complex problems. In this paper, we argue that *complexification*, i.e. the incremental elaboration of solutions through adding new structure, achieves both these goals. We demonstrate the power of complexification through the NeuroEvolution of Augmenting Topologies (NEAT) method, which evolves increasingly complex neural network architectures. NEAT is applied to an open-ended coevolutionary robot duel domain where robot controllers compete head to head. Because the robot duel domain supports a wide range of strategies, and because coevolution benefits from an escalating arms race, it serves as a suitable testbed for studying complexification. When compared to the evolution of networks with fixed structure, complexifying evolution discovers significantly more sophisticated strategies. The results suggest that in order to discover and improve complex solutions, evolution, and search in general, should be allowed to complexify as well as optimize.

## 1. Introduction

Evolutionary Computation (EC) is a class of algorithms that can be applied to open-ended learning problems in Artificial Intelligence. Traditionally, such algorithms evolve fixed-length genomes under the assumption that the space of the genome is sufficient to encode the solution. A genome containing $n$ genes encodes a single point in an $n$-dimensional search space. In many cases, a solution is known to exist somewhere in that space. For example, the global maximum of a function of three arguments *must* exist in the three dimensional space defined by those arguments. Thus, a genome of three genes can encode the location of the maximum.

However, many common structures are defined by an indefinite number of parameters. In particular, those solution types that can contain a variable number of parts can be represented by any number of parameters above some minimum. For example, the number of parts in neural networks, cellular automata, and electronic circuits can vary (Miller, Job, & Vassilev, 2000a; Mitchell, Crutchfield, & Das, 1996; Stanley & Miikkulainen, 2002d). In fact, theoretically two neural networks with different numbers of connections and nodes can represent the same function (Cybenko, 1989). Thus, it is not clear what number of genes is appropriate for solving a particular problem. Researchers evolving fixed-length genotypes must use heuristics to estimate *a priori* the appropriate number of genes to encode such structures.

A major obstacle to using fixed-length encodings is that heuristically determining the appropriate number of genes becomes impossible for very complex problems. For example, how many nodes and connections are necessary for a neural network that controls a ping-pong playing robot? Or, how many bits are needed in the neighborhood function of a cellular automaton that performs information compression? The answers to these questions can hardly be based on empirical experience or analytic methods, since little is known about the solutions. One possible approach is to simply make the genome extremely large, so that the space it encodes is extremely large and a solution is likely to lie somewhere within. Yet the larger the genome, the higher dimensional the space that evolution needs to search. Even if a ping-pong playing robot lies somewhere in the 10,000 dimensional space of a 10,000 gene genome, searching such a space may take prohibitively long.

Even more problematic are open-ended problems where phenotypes are meant to improve indefinitely and there is no known final solution. For example, in competitive games, estimating the complexity of the "best" possible player is difficult because such an estimate implicitly assumes that no better player can exist, which we cannot always know. Moreover, many artificial life domains are aimed at evolving increasingly complex artificial creatures for as long as possible (Maley, 1999). Such continual evolution is difficult with a fixed genome for two reasons: (1) When a good strategy is found in a fixed-length genome, the entire representational space of the genome is used to encode it. Thus, the only way to improve it is to *alter* the strategy, thereby sacrificing some of the functionality learned over previous generations. (2) Fixing the size of the genome in such domains arbitrarily fixes the maximum complexity of evolved creatures, defeating the purpose of the experiment.

In this paper, we argue that structured phenotypes can be evolved effectively by starting evolution with a population of small, simple genomes and systematically elaborating on them over generations by adding new genes. Each new gene expands the search space, adding a new dimension that previously did not exist. That way, evolution begins searching in a small easily-optimized space, and adds new dimensions as necessary. This approach is more likely to discover highly complex phenotypes than an approach that begins searching directly in the intractably large space of complete solutions. In fact, natural evolution utilizes this strategy, occasionally adding new genes that lead to increased phenotypic complexity (Martin 1999; Section 2). In biology, this process of incremental elaboration is called *complexification*, which is why we use this term to describe our approach as well.

In evolutionary computation, complexification refers to expanding the dimensionality of the search space while preserving the values of the majority of dimensions. In other words, complexification *elaborates* on the existing strategy by adding new structure without changing the existing representation. Thus the strategy does not only become different, but the number of possible responses to situations it can generate increases (Figure 1).

In the EC domain of neuroevolution (i.e. evolving neural networks), complexification means adding nodes and connections to already-functioning neural networks. This is the main idea behind NEAT (NeuroEvolution of Augmenting Topologies; Stanley and Miikkulainen 2002b,c,d), the method described in this paper. NEAT begins by evolving networks without any hidden nodes. Over many generations, new hidden nodes and connections are added, complexifying the space of potential solutions. In this way, more complex strategies elaborate on simpler strategies, focusing search on solutions that are likely to maintain existing capabilities.

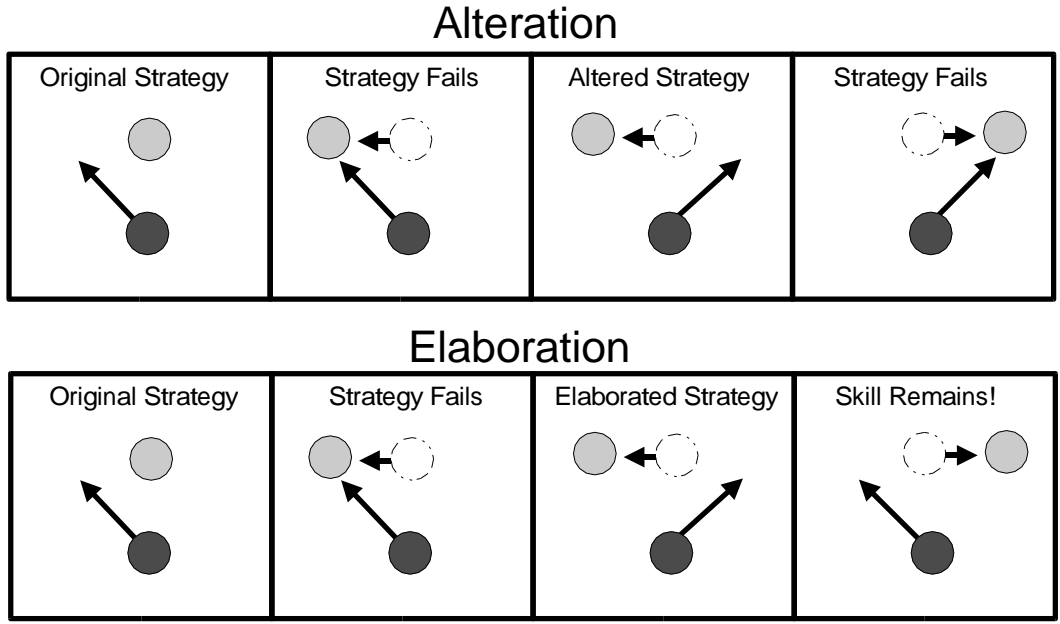

Figure 1: **Alteration vs. elaboration example.** The dark robot must evolve to avoid the lighter robot, which attempts to cause a collision. In the alteration scenario (top), the dark robot first evolves a strategy to go around the left side of the opponent. However, the strategy fails in a future generation when the opponent begins moving to the left. Thus, the dark robot alters its strategy by evolving the tendency to move right instead of left. However, when the light robot later moves right, the new, altered, strategy fails because the dark robot did not retain its old ability to move left. In the elaboration scenario (bottom), the original strategy of moving left also fails. However, instead of altering the strategy, it is *elaborated* by adding a new ability to move right as well. Thus, when the opponent later moves right, the dark robot still has the ability to avoid it by using its original strategy. Elaboration is necessary for a coevolutionary arms race to emerge and it can be achieved through complexification.

Expanding the length of the size of the genome has been found effective in previous work (Cliff, Harvey, & Husbands, 1993; Harvey, 1993; Koza, 1995; Lindgren & Johansson, 2001). NEAT advances this idea by making it possible to search a wide range of increasingly complex network topologies simultaneously. This process is based on three technical components: (1) Keeping track of which genes match up with which among differently sized genomes throughout evolution; (2) speciating the population so that solutions of differing complexity can exist independently; and (3) starting evolution with a uniform population of small networks. These components work together in complexifying solutions as part of the evolutionary process. In prior work, NEAT has been shown to solve challenging reinforcement learning problems more efficiently than other neuroevolution methods (Stanley and Miikkulainen 2002b,c,d). The focus of these studies was on optimizing a given fitness function through complexifying evolution.

In this paper, we focus on open-ended problems that have no explicit fitness function; instead, fitness depends on comparisons with other agents that are also evolving. The goal is to discover creative solutions beyond a designer's ability to define a fitness function. It is difficult to continually improve solutions in such *coevolutionary* domains because evolution tends to oscillate between idiosyncratic yet uninteresting solutions (Floreano & Nolfi, 1997). Complexification encourages continuing innovation by elaborating on existing solutions.

In order to demonstrate the power of complexification in coevolution, NEAT is applied to the competitive robot control domain of *Robot Duel*. There is no known optimal strategy in the domain but there is substantial room to come up with increasingly sophisticated strategies. The main results were that (1) evolution did complexify when possible, (2) complexification led to elaboration, and (3) significantly more sophisticated and successful strategies were evolved with complexification than without it. These results imply that complexification allows establishing a coevolutionary arms race that achieves a significantly higher level of sophistication than is otherwise possible.

We begin by reviewing biological support for complexification, as well as past work in coevolution, followed by a description of the NEAT method, and experimental results.

## 2. Background

The natural process of complexification has led to important biological innovations. Its most natural application in EC is in competitive coevolution, as will be reviewed below.

### 2.1 Complexification in Nature

Mutation in nature not only results in optimizing existing structures: New genes are occasionally added to the genome, allowing evolution to perform a complexifying function over and above optimization. In addition, complexification is protected in nature in that interspecies mating is prohibited. Such speciation creates important dynamics differing from standard GAs. In this section, we discuss these characteristics of natural evolution as a basis for our approach to utilizing them computationally in genetic algorithms.

*Gene duplication* is a special kind of mutation in which one or more parental genes are copied into an offspring's genome more than once. The offspring then has redundant genes expressing the same proteins. Gene duplication has been responsible for key innovations in overall body morphology over the course of natural evolution (Amores, Force, Yan, Joly, Amemiya, Fritz, Ho, Langeland, Prince, Wang, Westerfield, Ekker, & Postlethwait, 1998; Carroll, 1995; Force, Lynch, Pickett, Amores, lin Yan, & Postlethwait, 1999; Martin, 1999).

A major gene duplication event occurred around the time that vertebrates separated from invertebrates. The evidence for this duplication centers around *HOX genes*, which determine the fate of cells along the anterior-posterior axis of embryos. HOX genes are crucial in shaping the overall pattern of development in embryos. In fact, differences in HOX gene regulation explain a great deal of the diversity among arthropods and tetrapods (Carroll, 1995). Invertebrates have a single HOX cluster while vertebrates have four, suggesting that cluster duplication significantly contributed to elaborations in vertebrate body-plans (Amores et al., 1998; Holland, Garcia-Fernandez, Williams, & Sidow, 1994; Nadeau & Sankoff, 1997; Postlethwait, Yan, Gates, Horne, Amores, Brownlie, & Donovan, 1998; Sidow, 1996). The additional HOX genes took on new roles in regulating how vertebrate anterior-posterior axis develops, considerably increasing body-plan complexity. Although Martin (1999) argues that the additional clusters can be explained by many single gene duplications accumulating over generations, as opposed to massive whole-genome duplications, researchers agree that gene duplication in some form contributed significantly to body-plan elaboration.

A detailed account of how duplicate genes can take on novel roles was given by Force et al. (1999): Base pair mutations in the generations following duplication *partition* the initially redundant regulatory roles of genes into separate classes. Thus, the embryo develops in the same way, but the genes that determine the overall body-plan are confined to more specific roles, since there are more of them. The partitioning phase completes when redundant clusters of genes are separated enough so that they no longer produce identical proteins at the same time. After partitioning, mutations within the duplicated cluster of genes affect different steps in development than mutations within the original cluster. In other words, duplication creates more points at which mutations can occur. In this manner, developmental processes complexify.

Gene duplication is a possible explanation how natural evolution indeed expanded the size of genomes throughout evolution, and provides inspiration for adding new genes to artificial genomes as well. In fact, gene duplication motivated Koza (1995) to allow entire functions in genetic programs to be duplicated through a single mutation, and later differentiated through further mutations. When evolving neural networks, this process means adding new neurons and connections to the networks.

In order to implement this idea in artificial evolutionary systems, we are faced with two major challenges. First, such systems evolve differently sized and shaped network topologies, which can be difficult to cross over without losing information. For example, depending on when new structure was added, the same gene may exist at different positions, or conversely, different genes may exist at the same position. Thus, artificial crossover may disrupt evolved topologies through misalignment. Second, with variable-length genomes, it may be difficult to find innovative solutions. Optimizing many genes takes longer than optimizing only a few, meaning that more complex networks may be eliminated from the population before they have a sufficient opportunity to be optimized.

However, biological evolution also operates on variable-length genomes, and these problems did not stop complexification in nature. How are these problems avoided in biological evolution? First, nature has a mechanism for aligning genes with their counterparts during crossover, so that data is not lost nor obscured. This alignment process has been most clearly observed in *E. coli* (Radding, 1982; Sigal & Alberts, 1972). A special protein called *RecA* takes a single strand of DNA and aligns it with another strand at genes that express the same traits, which are called *homologous genes*. This process is called *synapsis*.

Second, innovations in nature are protected through speciation. Organisms with significantly divergent genomes never mate because they are in different species. If any organism could mate with any other, organisms with initially larger, less-fit genomes would be forced to compete for mates with their simpler, more fit counterparts. As a result, the larger, more innovative genomes would fail to produce offspring and disappear from the population. In contrast, in a speciated population, organisms with larger genomes compete for mates among their own species, instead of with the population at large. That way, organisms that may initially have lower fitness than the general population still have a chance to reproduce, giving novel concepts a chance to realize their potential without being prematurely eliminated. Because speciation benefits the evolution of diverse populations, a variety of speciation methods have been employed in EC (Goldberg & Richardson, 1987; Mahfoud, 1995; Ryan, 1994).

It turns out complexification is also possible in evolutionary computation if abstractions of synapsis and speciation are made part of the genetic algorithm. The NEAT method (section 3) is an implementation of this idea: The genome is complexified by adding new genes which in turn encode new structure in the phenotype, as in biological evolution.

Complexification is especially powerful in open-ended domains where the goal is to continually generate more sophisticated strategies. Competitive coevolution is a particularly important such domain, as will be reviewed in the next section.

## 2.2 Competitive Coevolution

In competitive coevolution, individual fitness is evaluated through competition with other individuals in the population, rather than through an absolute fitness measure. In other words, fitness signifies only the relative strengths of solutions; an increased fitness in one solution leads to a decreased fitness for another. Ideally, competing solutions will continually outdo one another, leading to an "arms race" of increasingly better solutions (Dawkins & Krebs, 1979; Rosin, 1997; Van Valin, 1973). Competitive coevolution has traditionally been used in two kinds of problems. First, it can be used to evolve interactive behaviors that are difficult to evolve in terms of an absolute fitness function. For example, Sims (1994) evolved simulated 3D creatures that attempted to capture a ball before an opponent did, resulting in a variety of effective interactive strategies. Second, coevolution can be used to gain insight into the dynamics of game-theoretic problems. For example, Lindgren & Johansson (2001) coevolved iterated Prisoner's Dilemma strategies in order to demonstrate how they correspond to stages in natural evolution.

Whatever the goal of a competitive coevolution experiment, interesting strategies will only evolve if the arms race continues for a significant number of generations. In practice, it is difficult to establish such an arms race. Evolution tends to find the simplest solutions that can win, meaning that strategies can switch back and forth between different idiosyncratic yet uninteresting variations (Darwen, 1996; Floreano & Nolfi, 1997; Rosin & Belew, 1997). Several methods have been developed to encourage the arms race (Angeline & Pollack, 1993; Ficici & Pollack, 2001; Noble & Watson, 2001; Rosin & Belew, 1997). For example, a "hall of fame" or a collection of past good strategies can be used to ensure that current strategies remain competitive against earlier strategies. Recently, Ficici and Pollack (2001) and Noble and Watson (2001) introduced a promising method called *Pareto coevolution*, which finds the best learners and the best teachers in two populations by casting coevolution as a multiobjective optimization problem. This information enables choosing the best individuals to reproduce, as well as maintaining an informative and diverse set of opponents.

Although such techniques allow sustaining the arms race longer, they do not directly encourage *continual coevolution*, i.e. creating new solutions that maintain existing capabilities. For example, no matter how well selection is performed, or how well competitors are chosen, if the search space is fixed, a limit will eventually be reached. Also, it may occasionally be easier to escape a local optimum by adding a new dimension to the search space than by searching for a new path through the original space.

For these reasons, complexification is a natural technique for establishing a coevolutionary arms race. Complexification elaborates strategies by adding new dimensions to the search space. Thus, progress can be made indefinitely long: Even if a global optimum is reached in the search space of solutions, new dimensions can be added, opening up a higher-dimensional space where even better optima may exist.

To test this idea experimentally, we chose a robot duel domain that combines predator/prey interaction and food foraging in a novel head-to-head competition (Section 4). We use this domain to demonstrate how NEAT uses complexification to continually elaborate solutions. The next section reviews the NEAT neuroevolution method, followed by a description of the robot duel domain and a discussion of the results.

## 3. NeuroEvolution of Augmenting Topologies (NEAT)

The NEAT method of evolving artificial neural networks combines the usual search for appropriate network weights with complexification of the network structure. This approach is highly effective, as shown e.g. in comparison to other neuroevolution (NE) methods in the double pole balancing benchmark task (Stanley & Miikkulainen, 2002b,c,d). The NEAT method consists of solutions to three fundamental challenges in evolving neural network topology: (1) What kind of genetic representation would allow disparate topologies to crossover in a meaningful way? Our solution is to use historical markings to line up genes with the same origin. (2) How can topological innovation that needs a few generations to optimize be protected so that it does not disappear from the population prematurely? Our solution is to separate each innovation into a different species. (3) How can topologies be minimized *throughout evolution* so the most efficient solutions will be discovered? Our solution is to start from a minimal structure and add nodes and connections incrementally. In this section, we explain how each of these solutions is implemented in NEAT.

### 3.1 Genetic Encoding

Evolving structure requires a flexible genetic encoding. In order to allow structures to complexify, their representations must be dynamic and expandable. Each genome in NEAT includes a list of *connection genes*, each of which refers to two *node genes* being connected. (Figure 2). Each connection gene specifies the in-node, the out-node, the weight of the connection, whether or not the connection gene is expressed (an enable bit), and an *innovation number*, which allows finding corresponding genes during crossover.

Mutation in NEAT can change both connection weights and network structures. Connection weights mutate as in any NE system, with each connection either perturbed or not. Structural mutations, which form the basis of complexification, occur in two ways (Figure 3). Each mutation expands the size of the genome by adding gene(s). In the *add connection* mutation, a single new connection gene is added connecting two previously unconnected nodes. In the *add node* mutation, an existing connection is split and the new node placed where the old connection used to be. The old connection is disabled and two new connections are added to the genome. The connection between the first node in the chain and the new node is given a weight of one, and the connection between the new node and the last node in the chain is given the same weight as the connection being split. Splitting the connection in this way introduces a nonlinearity (i.e. sigmoid function) where there was none before. When initialized in this way, this nonlinearity changes the function only slightly, and the new node is immediately integrated into the network. Old behaviors encoded in the preexisting network structure are not destroyed and remain qualitatively the same, while the new structure provides an opportunity to elaborate on these original behaviors.

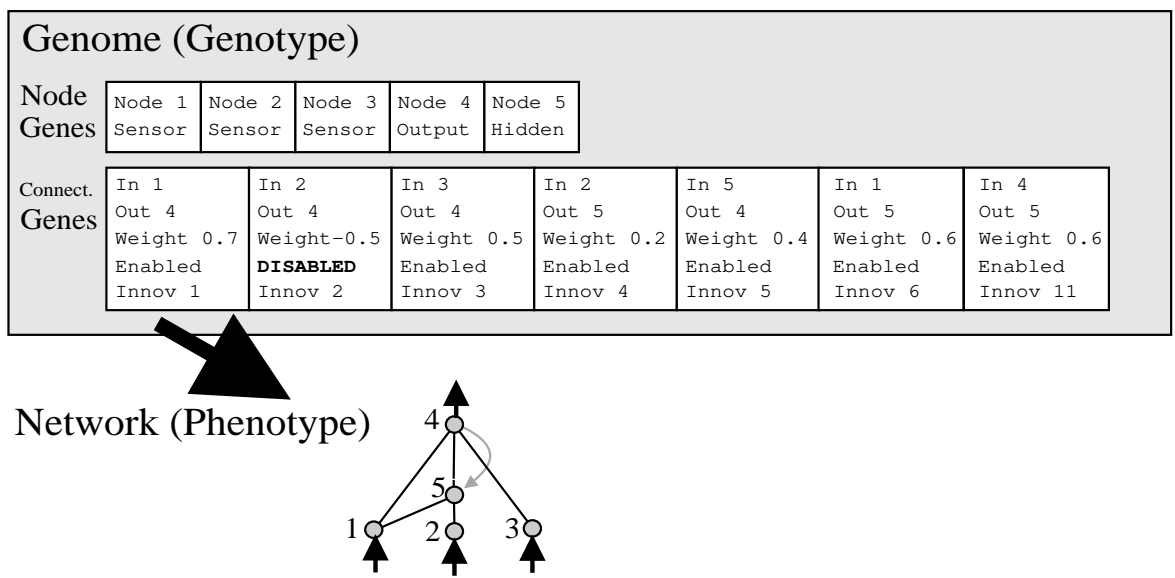

Figure 2: **A NEAT genotype to phenotype mapping example.** A genotype is depicted that produces the shown phenotype. There are 3 input nodes, one hidden, and one output node, and seven connection definitions, one of which is recurrent. The second gene is disabled, so the connection that it specifies (between nodes 2 and 4) is not expressed in the phenotype. In order to allow complexification, genome length is unbounded.

Through mutation, the genomes in NEAT will gradually get larger. Genomes of varying sizes will result, sometimes with different connections at the same positions. Crossover must be able to recombine networks with differing topologies, which can be difficult (Radcliffe, 1993). The next section explains how NEAT addresses this problem.

## 3.2 Tracking Genes through Historical Markings

It turns out that the historical origin of each gene can be used to tell us exactly which genes match up between *any* individuals in the population. Two genes with the same historical origin represent the same structure (although possibly with different weights), since they were both derived from the same ancestral gene at some point in the past. Thus, all a system needs to do is to keep track of the historical origin of every gene in the system.

Tracking the historical origins requires very little computation. Whenever a new gene appears (through structural mutation), a *global innovation number* is incremented and assigned to that gene. The innovation numbers thus represent a chronology of every gene in the system. As an example, let us say the two mutations in Figure 3 occurred one after another in the system. The new connection gene created in the first mutation is assigned the number 7, and the two new connection genes added during the new node mutation are assigned the numbers 8 and 9. In the future, whenever these genomes crossover, the offspring will inherit the same innovation numbers on each gene. Thus, the historical origin of every gene in the system is known throughout evolution.

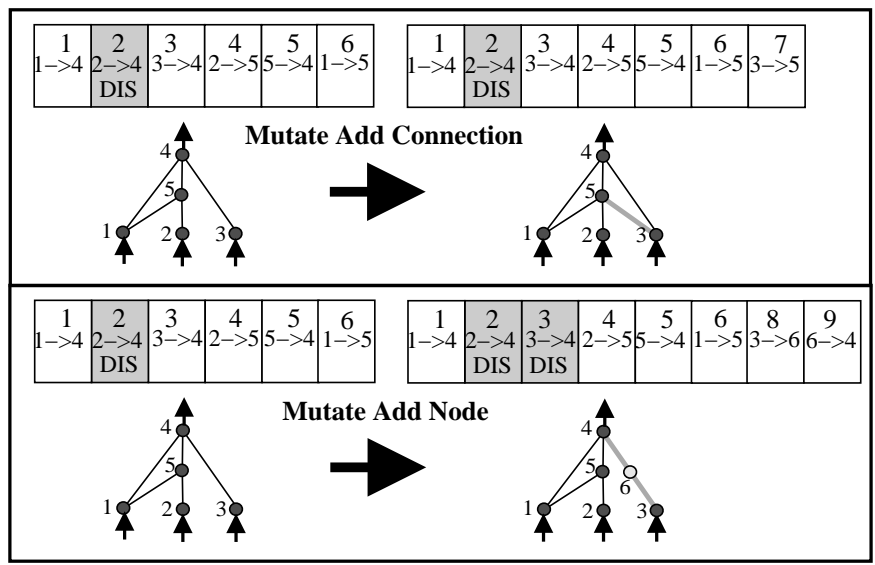

Figure 3: **The two types of structural mutation in NEAT.** Both types, adding a connection and adding a node, are illustrated with the genes above their phenotypes. The top number in each genome is the *innovation number* of that gene. The bottom two numbers denote the two nodes connected by that gene. The weight of the connection, also encoded in the gene, is not shown. The symbol *DIS* means that the gene is disabled, and therefore not expressed in the network. The figure shows how connection genes are appended to the genome when a new connection and a new node is added to the network. Assuming the depicted mutations occurred one after the other, the genes would be assigned increasing innovation numbers as the figure illustrates, thereby allowing NEAT to keep an implicit history of the origin of every gene in the population.

A possible problem is that the same structural innovation will receive different innovation numbers in the same generation if it occurs by chance more than once. However, by keeping a list of the innovations that occurred in the current generation, it is possible to ensure that when the same structure arises more than once through independent mutations in the same generation, each identical mutation is assigned the same innovation number. Through extensive experimentation, we established that resetting the list every generation as opposed to keeping a growing list of mutations throughout evolution is sufficient to prevent an explosion of innovation numbers.

Through innovation numbers, the system now knows exactly which genes match up with which (Figure 4). Genes that do not match are either *disjoint* or *excess*, depending on whether they occur within or outside the range of the other parent's innovation numbers. When crossing over, the genes with the same innovation numbers are lined up and are randomly chosen for the offspring genome. Genes that do not match are inherited from the more fit parent, or if they are equally fit, from both parents randomly. Disabled genes have a 25% chance of being reenabled during crossover, allowing networks to make use of older genes once again.[1]

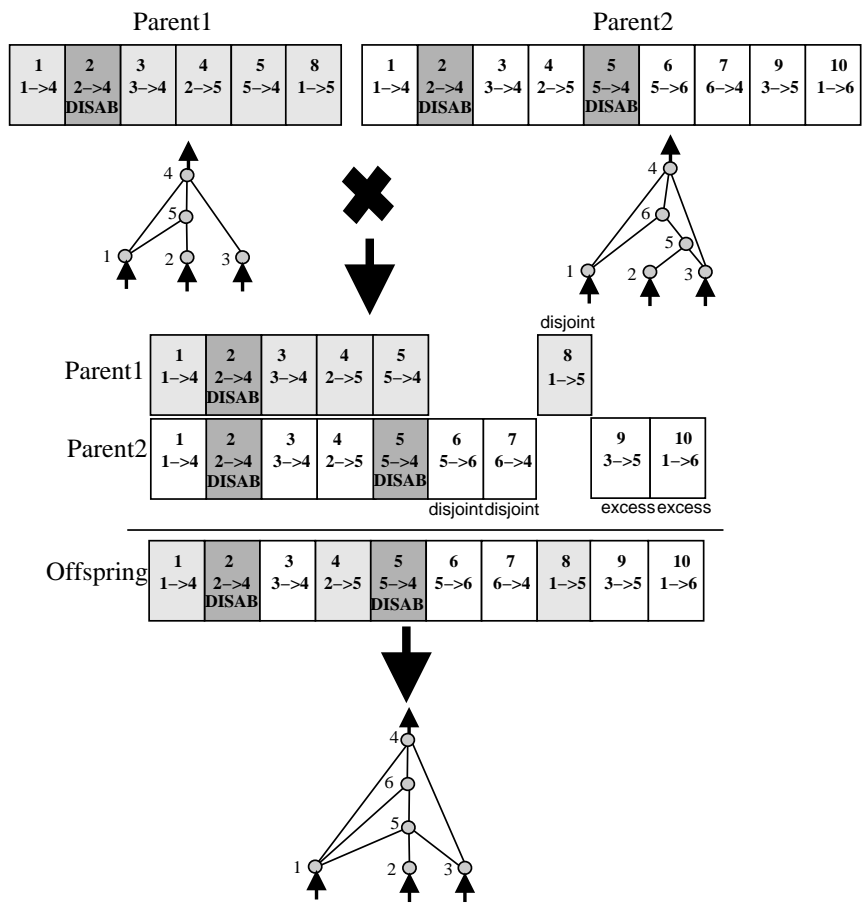

Figure 4: **Matching up genomes for different network topologies using innovation numbers.** Although Parent 1 and Parent 2 look different, their innovation numbers (shown at the top of each gene) tell us that several of their genes match up even without topological analysis. A new structure that combines the overlapping parts of the two parents as well as their different parts can be created in crossover. In this case, equal fitnesses are assumed, so the disjoint and excess genes from both parents are inherited randomly. Otherwise they would be inherited from the more fit parent. The disabled genes may become enabled again in future generations: There is a preset chance that an inherited gene is disabled if it is disabled in either parent.

Historical markings allow NEAT to perform crossover without the need for expensive topological analysis. Genomes of different organizations and sizes stay compatible throughout evolution, and the problem of comparing different topologies is essentially avoided. This methodology allows NEAT to complexify structure while still maintaining genetic compatibility. However, it turns out that a population of varying complexities cannot maintain topological innovations on its own. Because smaller structures optimize faster than larger structures, and adding nodes and connections usually initially decreases the fitness of the network, recently augmented structures have little hope of surviving more than one generation even though the innovations they represent might be crucial towards solving the task in the long run. The solution is to protect innovation by speciating the population, as explained in the next section.

---

1. See Appendix A for specific mating parameters used in the experiments in this paper.

### 3.3 Protecting Innovation through Speciation

NEAT speciates the population so that individuals compete primarily within their own niches instead of with the population at large. This way, topological innovations are protected and have time to optimize their structure before they have to compete with other niches in the population. In addition, speciation prevents bloating of genomes: Species with smaller genomes survive as long as their fitness is competitive, ensuring that small networks are not replaced by larger ones unnecessarily. Protecting innovation through speciation follows the philosophy that new ideas must be given time to reach their potential before they are eliminated.

Historical markings make it possible for the system to divide the population into species based on topological similarity. We can measure the distance $\delta$ between two network encodings as a linear combination of the number of excess ($E$) and disjoint ($D$) genes, as well as the average weight differences of matching genes ($\overline{W}$):

$$\delta = \frac{c_1 E}{N} + \frac{c_2 D}{N} + c_3 \cdot \overline{W}. \tag{1}$$

The coefficients $c_1$, $c_2$, and $c_3$ adjust the importance of the three factors, and the factor $N$, the number of genes in the larger genome, normalizes for genome size ($N$ can be set to 1 if both genomes are small). Genomes are tested one at a time; if a genome's distance to a randomly chosen member of the species is less than $\delta_t$, a compatibility threshold, it is placed into this species. Each genome is placed into the first species from the *previous generation* where this condition is satisfied, so that no genome is in more than one species. If a genome is not compatible with any existing species, a new species is created. The problem of choosing the best value for $\delta_t$ can be avoided by making $\delta_t$ *dynamic*; that is, given a target number of species, the system can raise $\delta_t$ if there are too many species, and lower $\delta_t$ if there are too few.

As the reproduction mechanism for NEAT, we use *explicit fitness sharing* (Goldberg & Richardson, 1987), where organisms in the same species must share the fitness of their niche. Thus, a species cannot afford to become too big even if many of its organisms perform well. Therefore, any one species is unlikely to take over the entire population, which is crucial for speciated evolution to maintain topological diversity. The adjusted fitness $f_i'$ for organism $i$ is calculated according to its distance $\delta$ from every other organism $j$ in the population:

$$f_i' = \frac{f_i}{\sum_{j=1}^{n} \text{sh}(\delta(i,j))}. \tag{2}$$

The sharing function sh is set to 0 when distance $\delta(i,j)$ is above the threshold $\delta_t$; otherwise, $\text{sh}(\delta(i,j))$ is set to 1 (Spears, 1995). Thus, $\sum_{j=1}^{n} \text{sh}(\delta(i,j))$ reduces to the number of organisms in the same species as organism $i$. This reduction is natural since species are already clustered by compatibility using the threshold $\delta_t$. Every species is assigned a potentially different number of offspring in proportion to the sum of adjusted fitnesses $f_i'$ of its member organisms. Species reproduce by first eliminating the lowest performing members from the population. The entire population is then replaced by the offspring of the remaining organisms in each species.

The net effect of speciating the population is that structural innovation is protected. The final goal of the system, then, is to perform the search for a solution as efficiently as possible. This goal is achieved through complexification from a simple starting structure, as detailed in the next section.

### 3.4 Minimizing Dimensionality through Complexification

Unlike other systems that evolve network topologies and weights (Angeline, Saunders, & Pollack, 1993; Gruau, Whitley, & Pyeatt, 1996; Yao, 1999; Zhang & Muhlenbein, 1993), NEAT begins with a uniform population of simple networks with no hidden nodes, differing only in their initial random weights. Speciation protects new innovations, allowing topological diversity to be gradually introduced over evolution. Thus, because NEAT protects innovation using speciation, it can start in this manner, minimally, and grow new structure over generations.

New structure is introduced incrementally as structural mutations occur, and only those structures survive that are found to be useful through fitness evaluations. This way, NEAT searches through a minimal number of weight dimensions, significantly reducing the number of generations necessary to find a solution, and ensuring that networks become no more complex than necessary. This gradual production of increasingly complex structures constitutes *complexification*. In other words, NEAT searches for the optimal topology by incrementally complexifying existing structure.

In previous work, each of the three main components of NEAT (i.e. historical markings, speciation, and starting from minimal structure) were experimentally ablated in order to demonstrate how they contribute to performance (Stanley & Miikkulainen, 2002b,d). The ablation study demonstrated that all three components are interdependent and necessary to make NEAT work. In this paper, we further show how complexification establishes the arms race in competitive coevolution. The next section describes the experimental domain in which this result will be demonstrated.

## 4. The Robot Duel Domain

Our hypothesis is that the complexification process outlined above allows discovering more sophisticated strategies, i.e. strategies that are more effective, flexible, and general, and include more components and variations than do strategies obtained through search in a fixed space. To demonstrate this effect, we need a domain where it is possible to develop a wide range increasingly sophisticated strategies, and where sophistication can be readily measured. A coevolution domain is particularly appropriate because a sustained arms race should lead to increasing sophistication.

In choosing the domain, it is difficult to strike a balance between being able to evolve complex strategies and being able to analyze and understand them. Pursuit and evasion tasks have been utilized for this purpose in the past (Gomez & Miikkulainen, 1997; Jim & Giles, 2000; Miller & Cliff, 1994; Reggia, Schulz, Wilkinson, & Uriagereka, 2001; Sims, 1994), and can serve as a benchmark domain for complexifying coevolution as well. While past experiments evolved either a predator or a prey, an interesting coevolution task can be established if the agents are instead equal and engaged in a duel. To win, an agent must develop a strategy that outwits that of its opponent, utilizing structure in the environment.

In this paper we introduce such a duel domain, in which two simulated robots try to overpower each other (Figure 5). The two robots begin on opposite sides of a rectangular room facing away from each other. As the robots move, they lose energy in proportion to the amount of force they apply to their wheels. Although the robots never run out of energy (they are given enough to survive the entire competition), the robot with higher energy wins when it collides with its competitor. In addition, each robot has a sensor indicating the difference in energy between itself and the other robot. To keep their energies high, the robots can consume food items, arranged in a symmetrical pattern in the room.

The robot duel task supports a broad range of sophisticated strategies that are easy to observe and interpret. The competitors must become proficient at foraging, prey capture, and escaping predators. In addition, they must be able to quickly switch from one behavior to another. The task is well-suited to competitive coevolution because naive strategies such as forage-then-attack can be complexified into more sophisticated strategies such as luring the opponent to waste its energy before attacking.

The simulated robots are similar to Kheperas (Mondada et al. 1993; Figure 6). Each has two wheels controlled by separate motors. Five rangefinder sensors can sense food and another five can sense the other robot. Finally, each robot has an energy-difference sensor, and a single wall sensor.

The robots are controlled with neural networks evolved with NEAT. The networks receive all of the robot sensors as inputs, as well as a constant bias that NEAT can use to change the activation thresholds of neurons. They produce three motor outputs: Two to encode rotation either right or left, and a third to indicate forward motion power. These three values are then translated into forces to be applied to the left and right wheels of the robot.

The state $s_t$ of the world at time $t$ is defined by the positions of the robots and food, the energy levels of the robots, and the internal states (i.e. neural activation) of the robots' neural networks, including sensors, outputs, and hidden nodes. The subsequent state $s_{t+1}$ is determined by the outputs of the robots' neural network controllers, computed from the inputs (i.e. sensor values) in $s_t$ in one step of propagation through the network. The robots change their position in $s_{t+1}$ according to their neural network outputs as follows. The change in direction of motion is proportional to the difference between the left and right motor outputs. The robot drives forward a distance proportional to the forward output on a continuous board of size 600 by 600. The robot first makes half its turn, then moves forward, then completes the second half of its turn, so that the turning and forward motions are effectively combined. If the robot encounters food, it receives an energy boost, and the food disappears from the world. The loss of energy due to movement is computed as the sum of the turn angle and the forward motion, so that even turning in place takes energy. If the robots are within a distance of 20, a collision occurs and the robot with a higher energy wins (see Appendix B for the exact parameter values used).

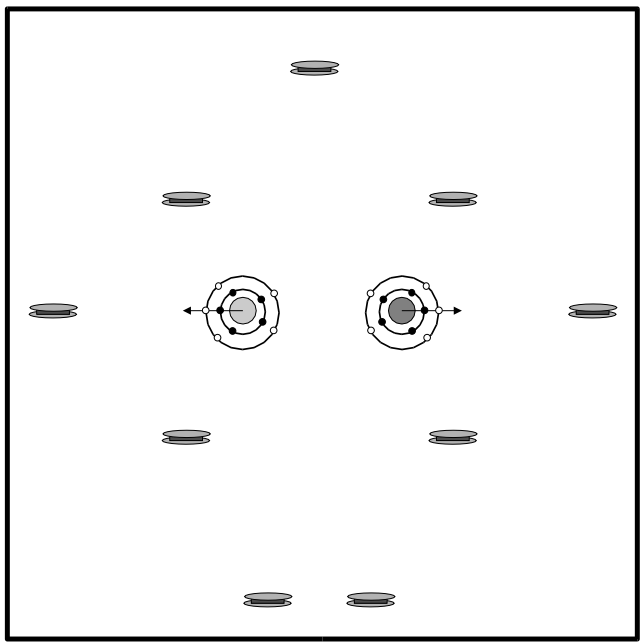

Figure 5: **The Robot Duel Domain.** The robots begin on opposite sides of the board facing away from each other as shown by the arrows pointing away from their centers. The concentric circles around each robot represent the separate rings of opponent sensors and food sensors available to each robot. Each ring contains five sensors. The robots lose energy when they move around, and gain energy by consuming food (shown as small sandwiches). The food is always placed in the depicted horizontally symmetrical pattern around the middle of the board. The objective is to attain a higher level of energy than the opponent, and then collide with it. Because of the complex interaction between foraging, pursuit, and evasion behaviors, the domain allows for a broad range of strategies of varying sophistication. Animated demos of such evolved strategies are available at **www.cs.utexas.edu/users/nn/pages/research/neatdemo.html**.

Since the observed state $o_t$ taken by the sensors does not include the internal state of the opponent robot, the robot duel is a partially-observable Markov decision process (POMDP). Since the next observed state $o_{t+1}$ depends on the decision of the opponent, it is necessary for robots to learn to predict what the opponent is likely to do, based on their past behavior, and also based on what is reasonable behavior in general. For example, it is reasonable to assume that if the opponent robot is quickly approaching and has higher energy, it is probably trying to collide. Because an important and complex portion of $s$ is not observable, memory, and hence recurrent connections, are crucial for success.

This complex robot-control domain allows competitive coevolution to evolve increasingly sophisticated and complex strategies, and can be used to understand complexification, as will be described next.

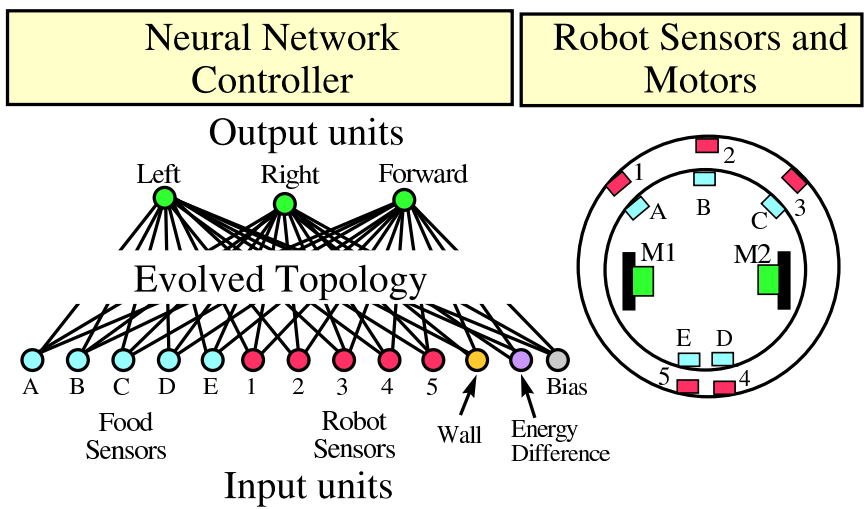

Figure 6: **Robot Neural Networks.** Five food sensors and five robot sensors detect the presence of objects around the robot. A single wall sensor indicates proximity to walls, and the energy difference sensor tells the robot how its energy level differs from that of its opponent. This difference is important because the robot with lower energy loses if the robots collide. The three motor outputs are mapped to forces that control the left and the right wheel.

## 5. Experiments

In order to demonstrate how complexification enhances performance, we ran thirty-three 500-generation runs of coevolution in the robot duel domain. Thirteen of these runs were based on the full NEAT method. Complexification was turned off in the remaining 20 runs (although networks were still speciated based on weight differences), in order to see how complexification contributes evolving sophisticated strategies. The competitive coevolution setup is described first, followed by an overview of the dominance tournament method for monitoring progress.

### 5.1 Competitive Coevolution Setup

The robot duel domain supports highly sophisticated strategies. Thus, the question in such a domain is whether *continual coevolution* will take place, i.e. whether increasingly sophisticated strategies will appear over the course of evolution. The experiment has to be set up carefully for this process to emerge, and to be able to identify it when it does.

In competitive coevolution, every network should play a sufficient number of games to establish a good measure of fitness. To encourage interesting and sophisticated strategies, networks should play a diverse and high quality sample of possible opponents. One way to accomplish this goal is to evolve two separate populations. In each generation, each population is evaluated against an intelligently chosen sample of networks from the other population. The population currently being evaluated is called the *host* population, and the population from which opponents are chosen is called the *parasite* population (Rosin &

Belew, 1997). The parasites are chosen for their quality and diversity, making host/parasite evolution more efficient and more reliable than one based on random or round robin tournament.

In the experiment, a single fitness evaluation included two competitions, one for the east and one for the west starting position. That way, networks needed to implement general strategies for winning, independent of their starting positions. Host networks received a single fitness point for each win, and no points for losing. If a competition lasted 750 time steps with no winner, the host received zero points.

In selecting the parasites for fitness evaluation, good use can be made of the speciation and fitness sharing that already occur in NEAT. Each host was evaluated against the four highest species' champions. They are good opponents because they are the best of the best species, and they are guaranteed to be diverse because their distance must exceed the threshold $\delta_t$ (section 3.3). Another eight opponents were chosen randomly from a Hall of Fame composed of all generation champions (Rosin & Belew, 1997). The Hall of Fame ensures that existing abilities need to be maintained to obtain a high fitness. Each network was evaluated in 24 games (i.e. 12 opponents, 2 games each), which was found to be effective experimentally. Together speciation, fitness sharing, and Hall of Fame comprise an effective competitive coevolution methodology.

It should be noted that complexification does not depend on the particular coevolution methodology. For example Pareto coevolution (Ficici & Pollack, 2001; Noble & Watson, 2001) could have been used as well, and the advantages of complexification would be the same. However, Pareto coevolution requires every member of one population to play every member of the other, and the running time in this domain would have been prohibitively long.

In order to interpret experimental results, a method is needed for analyzing progress in competitive coevolution. The next section describes such a method.

## 5.2 Monitoring Progress in Competitive Coevolution

A competitive coevolution run returns a record of every generation champion from both populations. The question is, how can a sequence of increasingly sophisticated strategies be identified in this data, if one exists? This section describes the *dominance tournament* method for monitoring progress in competitive coevolution (Stanley & Miikkulainen, 2002a) that allows us to do that.

First we need a method for performing individual comparisons, i.e. whether one strategy is better than another. Because the board configurations can vary between games, champion networks were compared on 144 different food configurations from each side of the board, giving 288 total games for each comparison. The food configurations included the same 9 symmetrical food positions used during training, plus an additional 2 food items, which were placed in one of 12 different positions on the east and west halves of the board. Some starting food positions give an initial advantage to one robot or another, depending on how close they are to the robots' starting positions. Thus, the one who wins the majority of the 288 total games has demonstrated its superiority in many different scenarios, including those beginning with a disadvantage. We say that network $a$ *is superior* to network $b$ if $a$ wins more games than $b$ out of the 288 total games.

Given this definition of superiority, progress can be tracked. The obvious way to do it is to compare each network to others throughout evolution, finding out whether later strategies can beat more opponents than earlier strategies. For example, Floreano & Nolfi (1997) used a measure called *master tournament*, in which the champion of each generation is compared to all other generation champions. Unfortunately, such methods are impractical in a time-intensive domain such as the robot duel competition. Moreover, the master tournament only counts how many strategies can be defeated by each generation champion, without identifying which ones. Thus, it can fail to detect cases where strategies that defeat fewer previous champions are actually superior in a direct comparison. For example, if strategy $A$ defeats 499 out of 500 opponents, and $B$ defeats 498, the master tournament will designate $A$ as superior to $B$ *even if* $B$ defeats $A$ in a direct comparison. In order to decisively track strategic innovation, we need to identify *dominant strategies*, i.e. those that defeat *all previous* dominant strategies. This way, we can make sure that evolution proceeds by developing a progression of strictly more powerful strategies, instead of e.g. switching between alternative ones.

The *dominance tournament* method of tracking progress in competitive coevolution meets this goal (Stanley & Miikkulainen, 2002a). Let a *generation champion* be the winner of a 288 game comparison between the host and parasite champions of a single generation. Let $d_j$ be the $j$th dominant strategy to appear over evolution. Dominance is defined recursively starting from the first generation and progressing to the last:

- The first dominant strategy $d_1$ is the generation champion of the first generation;

- dominant strategy $d_j$, where $j > 1$, is a generation champion such that for all $i < j$, $d_j$ is superior to $d_i$ (i.e. wins the 288 game comparison with it).

This strict definition of dominance prohibits circularities. For example, $d_4$ must be superior to strategies $d_1$ through $d_3$, $d_3$ superior to both $d_1$ and $d_2$, and $d_2$ superior to $d_1$. We call $d_n$ the $n$th dominant strategy of the run. If a network $c$ exists that, for example, defeats $d_4$ but loses to $d_3$, making the superiority circular, it would not satisfy the second condition and would not be entered into the dominance hierarchy.

The entire process of deriving a dominance hierarchy from a population is a *dominance tournament*, where competitors play all previous dominant strategies until they either lose a 288 game comparison, or win every comparison to previous dominant strategies, thereby becoming a new dominant strategy. Dominance tournament allows us to identify a sequence of increasingly more sophisticated strategies. It also requires significantly fewer comparisons than the master tournament (Stanley & Miikkulainen, 2002a).

Armed with the appropriate coevolution methodology and a measure of success, we can now ask the question: Does the complexification result in more successful competitive coevolution?

## 6. Results

Each of the 33 evolution runs took between 5 and 10 days on a 1GHz Pentium III processor, depending on the progress of evolution and sizes of the networks involved. The NEAT algorithm itself used less than 1% of this computation: Most of the time was spent in evaluating networks in the robot duel task. Evolution of fully-connected topologies took about 90% longer than structure-growing NEAT because larger networks take longer to evaluate.

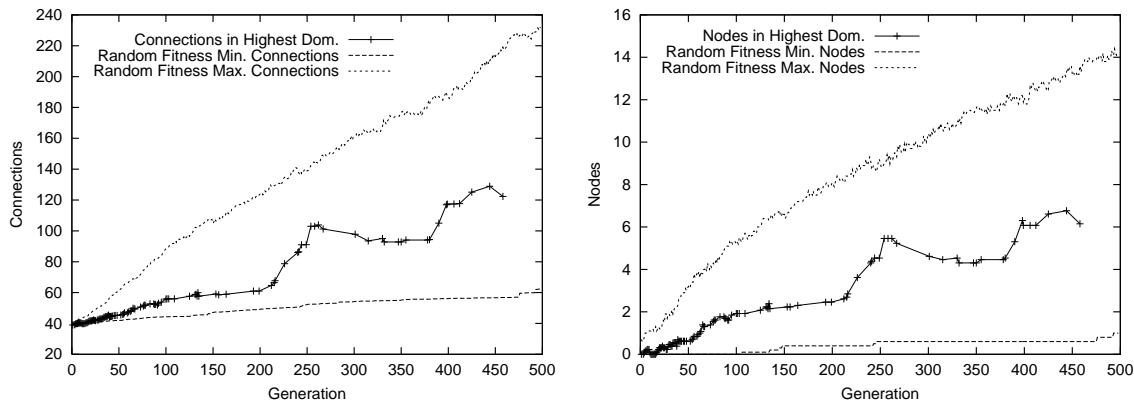

Figure 7: **Complexification of connections and nodes over generations.** The hashed lines depict the average number of connections and the average number of hidden nodes in the highest dominant network in each generation. Averages are taken over 13 complexifying runs. A hash mark appears every generation in which a new dominant strategy emerged in at least one of the 13 runs. The graphs show that as dominance increases, so does complexity. The differences between the average final and first dominant strategies are statistically significant for both connections and nodes ($p < 0.001$). For comparison the dashed lines depict the sizes of the average smallest and largest networks in the entire population over five runs where the fitness is assigned randomly. These bounds show that the increase in complexity is not inevitable; both very simple and very complex species exist in the population throughout the run. When the dominant networks complexify, they do so because it is beneficial.

In order to analyze the results, we define *complexity* as the number of nodes and connections in a network: The more nodes and connections there are in the network, the more complex behavior it can potentially implement. The results were analyzed to answer three questions: (1) As evolution progresses does it also continually complexify? (2) Does such complexification lead to more sophisticated strategies? (3) Does complexification allow better strategies to be discovered than does evolving fixed-topology networks? Each question is answered in turn below.

## 6.1 Evolution of Complexity

NEAT was run thirteen times, each time from a different seed, to verify that the results were consistent. The highest levels of dominance achieved were 17, 14, 17, 16, 16, 18, 19, 15, 17, 12, 12, 11, and 13, averaging at 15.15 (sd = 2.54).

At each generation where the dominance level increased in at least one of the thirteen runs, we averaged the number of connections and number of nodes in the current dominant strategy across all runs (Figure 7). Thus, the graphs represent a total of 197 dominance transitions spread over 500 generations. The rise in complexity is dramatic, with the average number of connections tripling and the average number of hidden nodes rising from 0 to almost six. In a smooth trend over the first 200 generations, the number of connections in the dominant strategy grows by 50%. During this early period, dominance transitions occur frequently (fewer prior strategies need to be beaten to achieve dominance). Over the next 300 generations, dominance transitions become more sparse, although they continue to occur.

Between the 200th and 500th generations a stepped pattern emerges: The complexity first rises dramatically, then settles, then abruptly increases again (This pattern is even more marked in individual complexifying runs; the averaging done in Figure 7 smooths it out somewhat). The cause for this pattern is speciation. While one species is adding a large amount of structure, other species are optimizing the weights of less complex networks. Initially, added complexity leads to better performance, but subsequent optimization takes longer in the new higher-dimensional space. Meanwhile, species with smaller topologies have a chance to temporarily catch up through optimizing their weights. Ultimately, however, more complex structures eventually win, since higher complexity is necessary for continued innovation.

Thus, there are two underlying forces of progress: The building of new structures, and the continual optimization of prior structures in the background. The product of these two trends is a gradual stepped progression towards increasing complexity.

An important question is: Because NEAT searches by adding structure only, not by removing it, does the complexity always increase whether it helps in finding good solutions or not? To demonstrate that NEAT indeed prefers simple solutions and complexifies only when it is useful, we ran five complexifying evolution runs with fitness assigned randomly (i.e. the winner of each game was chosen at random). As expected, NEAT kept a wide range of networks in its population, from very simple to highly complex (Figure 7). That is, the dominant networks did not *have to* become more complex; they only did so because it was beneficial. Not only is the minimum complexity in the random-fitness population much lower than that of the dominant strategies, but the maximum complexity is significantly greater. Thus, evolution complexifies sparingly, only when the complex species holds its own in comparison with the simpler ones.

## 6.2 Sophistication through Complexification

To see how complexification contributes to evolution, let us observe how a sample dominant strategy develops over time. While many complex networks evolved in the experiments, we follow the species that produced the winning network $d_{17}$ in the third run because its progress is rather typical and easy to understand. Let us use $S_k$ for the best network in species $S$ at generation $k$, and $h_l$ for the $l$th hidden node to arise from a structural mutation over the course of evolution. We will track both strategic and structural innovations in order to see how they correlate. Let us begin with $S_{100}$ (Figure 8a), when $S$ had a mature zero-hidden-node strategy:

- $S_{100}$'s main strategy was to follow the opponent, putting it in a position where it might by chance collide with its opponent when its energy is up. However, $S_{100}$ followed the opponent even when the opponent had more energy, leaving $S_{100}$ vulnerable to attack. $S_{100}$ did not clearly switch roles between foraging and chasing the enemy, causing it to miss opportunities to gather food.

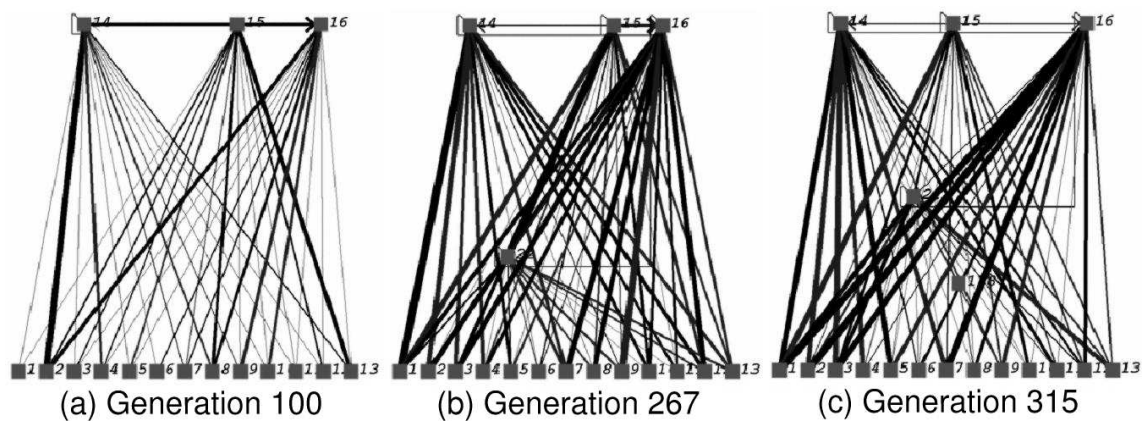

|                        |                        |                        |
| :--------------------: | :--------------------: | :--------------------: |
| (a) Generation 100     | (b) Generation 267     | (c) Generation 315     |

Figure 8: **Complexification of a Winning Species.** The best networks in the same species are depicted at landmark generations. Nodes are depicted as squares beside their node numbers, and line thickness represents the strength of connections. Over time, the networks became more complex and gained skills. (a) The champion from generation 10 had no hidden nodes. (b) The addition of $h_{22}$ and its respective connections gave new abilities. (c) The appearance of $h_{172}$ refined existing behaviors.

- $S_{200}$. During the next 100 generations, $S$ evolved a *resting* strategy, which it used when it had significantly lower energy than its opponent. In such a situation, the robot stopped moving, while the other robot wasted energy running around. By the time the opponent got close, its energy was often low enough to be attacked. The resting strategy is an example of improvement that can take place without complexification: It involved increasing the inhibition from the energy difference sensor, thereby slightly modifying intensity of an existing behavior.

- In $S_{267}$ (Figure 8b), a new hidden node, $h_{22}$, appeared. This node arrived through an interspecies mating, and had been optimized for several generations already. Node $h_{22}$ gave the robot the ability to change its behavior at once into an all-out attack. Because of this new skill, $S_{267}$ no longer needed to follow the enemy closely at all times, allowing it to collect more food. By implementing this new strategy through a new node, it was possible not to interfere with the already existing resting strategy, so that $S$ now switched roles between resting when at a disadvantage to attacking when high on energy. Thus, the new structure resulted in strategic elaboration.

- In $S_{315}$ (Figure 8c), another new hidden node, $h_{172}$, split a link between an input sensor and $h_{22}$. Replacing a direct connection with a sigmoid function greatly improved $S_{315}$'s ability to attack at appropriate times, leading to very accurate role switching between attacking and foraging. $S_{315}$ would try to follow the opponent from afar focusing on resting and foraging, and only zoom in for attack when victory was certain. This final structural addition shows how new structure can improve the accuracy and timing of existing behaviors.

The analysis above shows that in some cases, weight optimization alone can produce improved strategies. However, when those strategies need to be extended, adding new structure allows the new behaviors to coexist with old strategies. Also, in some cases it is necessary to add complexity to make the timing or execution of the behavior more accurate. These results show how complexification can be utilized to produce increasing sophistication.

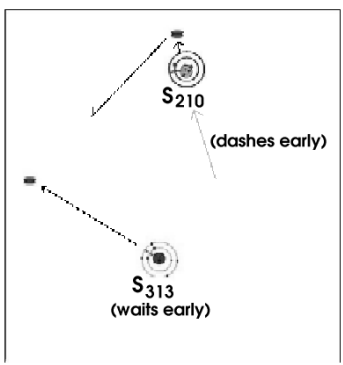

Figure 9: **Sophisticated Endgame.** Robot $S_{313}$ dashes for the last piece of food while $S_{210}$ is still collecting the second-to-last piece. Although it appeared that $S_{313}$ would lose because $S_{210}$ got the second-to-last piece, (gray arrow), it turns out that $S_{210}$ ends with a disadvantage. It has no chance to get to the last piece of food before $S_{313}$, and $S_{313}$ has been saving energy while $S_{210}$ wasted energy traveling long distances. This way, sophisticated strategies evolve through complexification, combining multiple objectives, and utilizing weaknesses in the opponent's strategy.

In order to illustrate the level of sophistication achieved in this process, we conclude this section by describing the competition between two sophisticated strategies, $S_{210}$ and $S_{313}$, from another run of evolution. At the beginning of the competition, $S_{210}$ and $S_{313}$ collected most of the available food until their energy levels were about equal. Two pieces of food remained on the board in locations distant from both $S_{210}$ and $S_{313}$ (Figure 9). Because of the danger of colliding with similar energy levels, the obvious strategy is to rush for the last two pieces of food. In fact, $S_{210}$ did exactly that, consuming the second-to-last piece, and then heading towards the last one. In contrast, $S_{313}$ forfeited the race for the second-to-last piece, opting to sit still, even though its energy temporarily dropped below $S_{210}$'s. However, $S_{313}$ was closer to the last piece and got there first. It received a boost of energy while $S_{210}$ wasted its energy running the long distance from the second-to-last piece. Thus, $S_{313}$'s strategy ensured that it had more energy when they finally met. Robot $S_{313}$'s behavior was surprisingly deceptive, showing that high strategic sophistication had evolved. Similar waiting behavior was observed against several other opponents, and also evolved in several other runs, suggesting that it was a robust result.

This analysis of individual evolved behaviors shows that complexification indeed elaborates on existing strategies, and allows highly sophisticated behaviors to develop that balance multiple goals and utilize weaknesses in the opponent. The last question is whether complexification indeed is necessary to achieve these behaviors.

## 6.3 Complexification vs. Fixed-topology Evolution and Simplification

Complexifying coevolution was compared to two alternatives: standard coevolution in a fixed search space, and to simplifying coevolution from a complex starting point. Note that it is not possible to compare methods using the standard crossvalidation techniques because no external performance measure exists in this domain. However, the evolved neural networks can be compared *directly* by playing a duel. Thus, for example, a run of fixed-topology coevolution can be compared to a run of complexifying coevolution by playing the highest dominant strategy from the fixed-topology run against the entire dominance ranking of the complexifying run. The highest level strategy in the ranking that the fixed-topology strategy can defeat, normalized by the number of dominance levels, is a measure of its quality against the complexifying coevolution. For example, if a strategy can defeat up to and including the 13th dominant strategy out of 15, then its performance against that run is $\frac{13}{15} = 86.7\%$. By playing every fixed-topology champion, every simplifying coevolution champion, and every complexifying coevolution champion against the dominance ranking from every complexifying run and averaging, we can measure the relative performance of each of these methods.

In this section, we will first establish the baseline performance by playing complexifying coevolution runs against themselves and demonstrating that a comparison with dominance levels is a meaningful measure of performance. We will then compare complexification with fixed-topology coevolution of networks of different architectures, including fully-connected small networks, fully-connected large networks, and networks with an optimal structure as determined by complexifying coevolution. Third, we will compare the performance of complexification with that of simplifying coevolution.

### 6.3.1 COMPLEXIFYING COEVOLUTION

The highest dominant strategy from each of the 13 complexifying runs played the entire dominance ranking from every other run. Their average performance scores were 87.9%, 83.8%, 91.9%, 79.4%, 91.9%, 99.6%, 99.4%, 99.5%, 81.8%, 96.2%, 90.6%, 96.9%, and 89.3%, with an overall average of 91.4% (sd=12.8%). Above all, this result shows that complexifying runs produce consistently good strategies: On average, the highest dominant strategies qualify for the top 10% of the other complexifying runs. The best runs were the sixth, seventh, and eighth, which were able to defeat almost the entire dominance ranking of every other run. The highest dominant network from the best run (99.6%) is shown in Figure 10.

In order to understand what it means for a network to be one or more dominance levels above another, Figure 11 shows how many more games the more dominant network can be expected to win on average over all its 288-game comparisons than the less dominant network. Even at the lowest difference (i.e. that of one dominance level), the more dominant network can be expected to win about 50 more games on average, showing that each difference in dominance level is important. The difference grows approximately linearly: A network 5 dominance levels higher will win 100 more games, while a network 10 levels higher will win 150 and 15 levels higher will win 200. These results suggest that dominance level comparisons indeed constitute a meaningful way to measure relative performance.

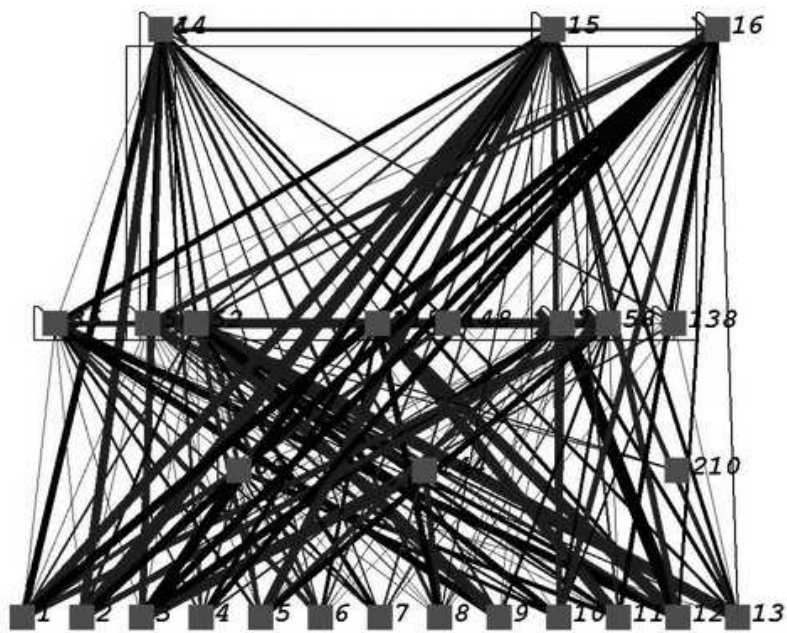

Figure 10: **The Best Complexifying Network.**
The highest dominant network from the sixth complexifying coevolution run was able to beat 99.6% of the dominance hierarchy of the other 12 runs. The network has 11 hidden units and 202 connections (plotted as in figure 8), making significant use of structure. While it still contains the basic direct connections, the strategy they represent has been elaborated by adding several new nodes and connections. For example, lateral and recurrent connections allow taking past events into account, resulting in more refined decisions. While such structures can be found reliably through complexification, it turns out difficult to discover them directly in the high dimensional space through fixed-topology evolution or through simplification.

### 6.3.2 FIXED-TOPOLOGY COEVOLUTION OF LARGE NETWORKS

In fixed-topology coevolution, the network architecture must be chosen by the experimenter. One sensible approach is to approximate the complexity of the best complexifying network. (Figure 10). This network included 11 hidden units and 202 connections, with both recurrent connections and direct connections from input to output. As an idealization of this structure, we used 10-hidden-unit fully recurrent networks with direct connections from inputs to outputs, with a total of 263 connections. A network of this type should be able to approximate the functionality of the most effective complexifying strategy. Fixed-topology coevolution runs exactly as complexifying coevolution in NEAT, except that no structural mutations can occur. In particular, the population is still speciated based on weight differences (i.e. $\overline{W}$ from equation 1), using the usual speciation procedure.

Three runs of fixed-topology coevolution were performed with these networks, and their highest dominant strategies were compared to the entire dominance ranking of every complexifying run. Their average performances were 29.1%, 34.4%, and 57.8%, for an overall average of 40.4%. Compared to the 91.4% performance of complexifying coevolution, it is clear that fixed-topology coevolution produced consistently inferior solutions. As a matter of fact, no fixed-topology run could defeat any of the highest dominant strategies from the 13 complexifying coevolution runs.

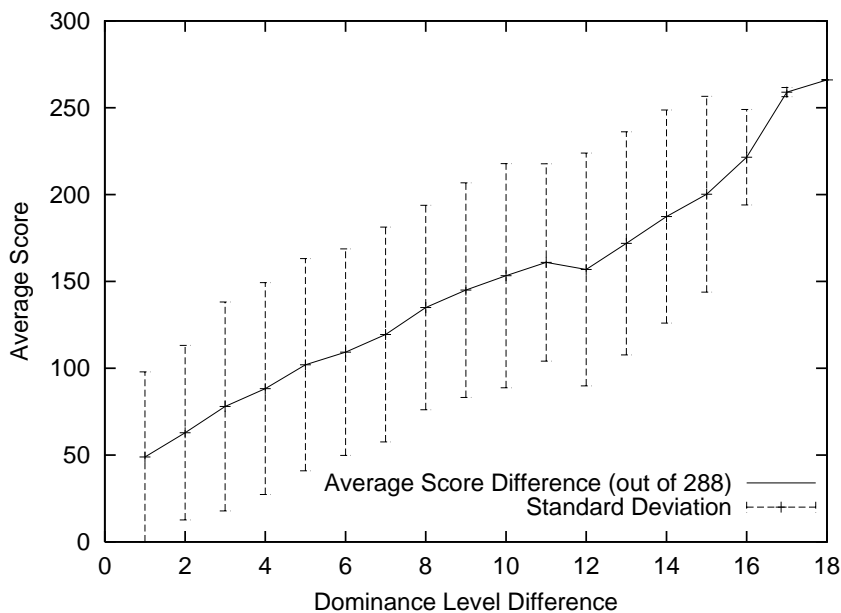

Figure 11: **Interpreting Differences in Dominance Levels.** The graph shows how many games in a 288-game comparison a more dominant network can be expected to win, averaged over all runs at all dominance levels of complexifying coevolution. For example, a network one level higher wins 50 more games out of 288. A larger difference in dominance levels translates to a larger difference in performance approximately linearly, suggesting that dominance levels can be used as a measure of performance when an absolute measure is not available.

This difference in performance can be illustrated by computing the *average generation* of complexifying coevolution with the same performance as fixed-topology coevolution. This generation turns out to be 24 (sd = 8.8). In other words, 500 generations of fixed-topology coevolution reach on average the same level of dominance as only 24 generations of complexifying coevolution! In effect, the progress of the entire fixed-topology coevolution run is compressed into the first few generations of complexifying coevolution (Figure 12).

### 6.3.3 FIXED-TOPOLOGY COEVOLUTION OF SMALL NETWORKS

One of the arguments for using complexifying coevolution is that starting the search directly in the space of the final solution may be intractable. This argument may explain why the attempt to evolve fixed-topology solutions at a high level of complexity failed. Thus, in the next experiment we aimed at reducing the search space by evolving fully-connected, fully-recurrent networks with a small number of hidden nodes as well as direct connections from inputs to outputs. After considerable experimentation, we found out that five hidden nodes (144 connections) was appropriate, allowing fixed-topology evolution to find the best solutions it could. Five hidden nodes is also about the same number of hidden nodes as the highest dominant strategies had on average in the complexifying runs.

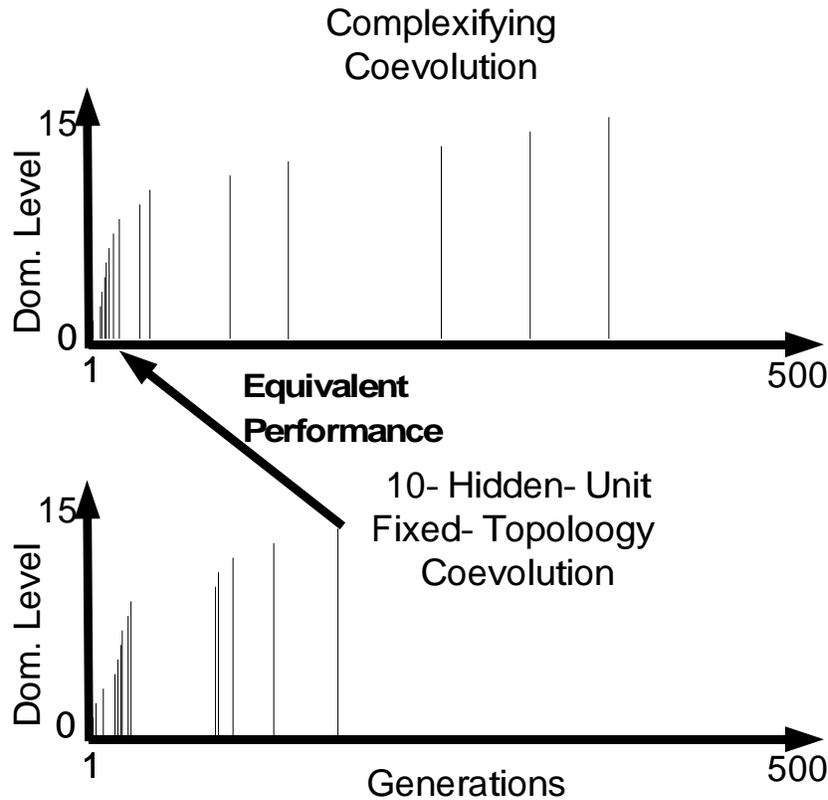

Figure 12: **Comparing Typical Runs of Complexifying Coevolution and Fixed-Topology Coevolution with Ten Hidden Units.** Dominance levels are charted on the $y$-axis and generations on the $x$-axis. A line appears at every generation where a new dominant strategy arose in each run. The height of the line represents the level of dominance. The arrow shows that the highest dominant strategy found in 10-hidden-unit fixed-topology evolution only performs as well as the 8th dominant strategy in the complexifying run, which was found in the 19th generation. (Average = 24, sd = 8.8) In other words, only a few generations of complexifying coevolution are as effective as several hundred of fixed-topology evolution.

A total of seven runs were performed with the 5-hidden-node networks, with average performances of 70.7%, 85.5%, 66.1%, 87.3%, 80.8%, 88.8%, and 83.1%. The overall average was 80.3% (sd=18.4%), which is better but still significantly below the 91.4% performance of complexifying coevolution ($p < 0.001$).

In particular, the two most effective complexifying runs were still never defeated by any of the fixed-topology runs. Also, because each dominance level is more difficult to achieve than the last, on average the fixed-topology evolution only reached the performance of the 159th complexifying generation (sd=72.0). Thus, even in the best case, fixed-topology coevolution on average only finds the level of sophistication that complexifying coevolution finds halfway through a run (Figure 13).

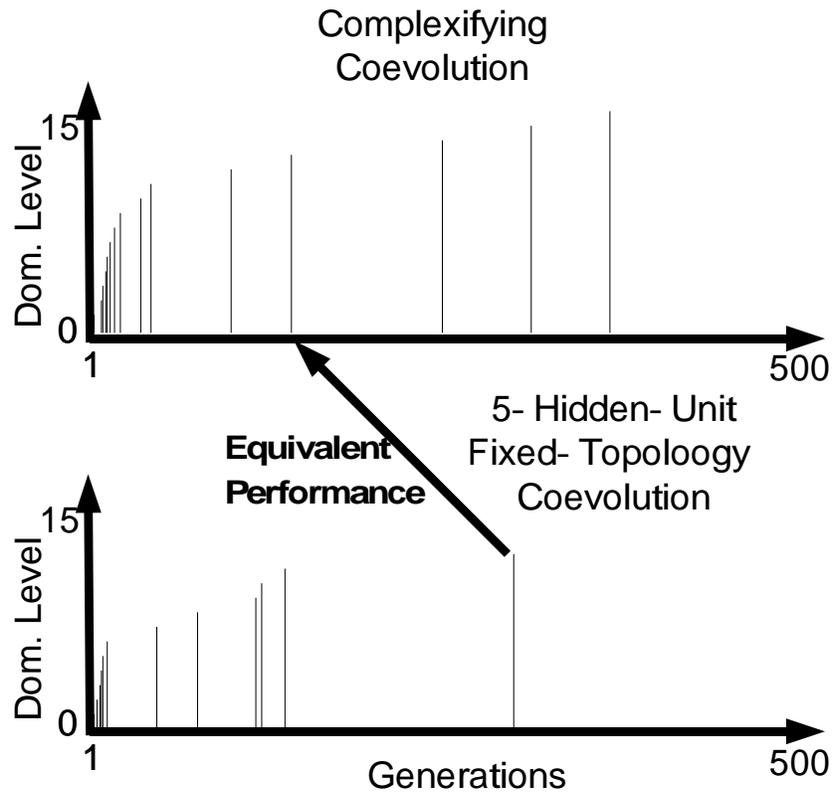

Figure 13: **Comparing Typical Runs of Complexifying Coevolution and Fixed-Topology Coevolution with Five Hidden Units.** As in figure 12, dominance levels are charted on the $y$-axis and generations on the $x$-axis, a line appears at every generation where a new dominant strategy arose in each run, and the height of the line represents the level of dominance. The arrow shows that the highest dominant strategy found in the 5-hidden-unit fixed-topology evolution only performs as well as the 12th dominant strategy in the complexifying run, which was found in the 140th generation (Average 159, sd = 72.0). Thus, even in the best configuration, fixed-topology evolution takes about twice as long to achieve the same level of performance.

### 6.3.4 FIXED-TOPOLOGY COEVOLUTION OF BEST COMPLEXIFYING NETWORK

One problem with evolving fully-connected architectures is that they may not have an appropriate topology for this domain. Of course, it is very difficult to guess an appropriate topology *a priori*. However, it is still interesting to ask whether fixed-topology coevolution could succeed in the task assuming that the right topology was known? To answer this question, we evolved networks as in the other fixed-topology experiments, except this time using the topology of the best complexifying network (Figure 10). This topology may constrain the search space in such a way that finding a sophisticated solution is more likely than with a fully-connected architecture. If so, it is possible that seeding the population with a successful topology gives it an advantage even over complexifying coevolution, which must build the topology from a minimal starting point.

Five runs were performed, obtaining average performance score 86.2%, 83.3%, 88.1%, 74.2%, and 80.3%, and an overall average of 82.4% (sd=15.1%). The 91.4% performance of complexifying coevolution is significantly better than even this version of fixed-topology coevolution ($p < 0.001$). However, interestingly, the 40.4% average performance of 10-hidden-unit fixed topology coevolution is significantly *below* best-topology evolution, even though both methods search in spaces of similar sizes. In fact, best-topology evolution performs at about the same level as 5-hidden-unit fixed-topology evolution (80.3%), even though 5-hidden-unit evolution optimizes half the number of hidden nodes. Thus, the results confirm the hypothesis that using a successful evolved topology does help constrain the search. However, in comparison to complexifying coevolution, the advantage gained from starting this way is still not enough to make up for the penalty of starting search directly in a high-dimensional space. As Figure 14 shows, best-topology evolution on average only finds a strategy that performs as well as those found by the 193rd generation of complexifying coevolution.

The results of the fixed-topology coevolution experiments can be summarized as follows: If this method is used to search directly in the high-dimensional space of the most effective solutions, it reaches only 40% of the performance of complexifying coevolution. It does better if it is allowed to optimize less complex networks; however, the most sophisticated solutions may not exist in that space. Even given a topology appropriate for the task, it does not reach the same level as complexifying coevolution. Thus, fixed-topology coevolution does not appear to be competitive with complexifying coevolution with *any* choice of topology.

The conclusion is that complexification is superior not only because it allows discovering the appropriate high-dimensional topology automatically, but also because it makes the optimization of that topology more efficient. This point will be discussed further in Section 7.

### 6.3.5 SIMPLIFYING COEVOLUTION

A possible remedy to having to search in high-dimensional spaces is to allow evolution to search for smaller structures by removing structure incrementally. This *simplifying coevolution* is the opposite of complexifying coevolution. The idea is that a mediocre complex solution can be refined by removing unnecessary dimensions from the search space, thereby accelerating the search.

Although simplifying coevolution is an alternative method to complexifying coevolution for finding topologies, it still requires a complex starting topology to be specified. This topology was chosen with two goals in mind: (1) Simplifying coevolution should start with sufficient complexity to at least potentially find solutions of equal or more complexity than the best solutions from complexifying coevolution, and (2) with a rate of structural removal equivalent to the rate of structural addition in complexifying NEAT, it should be possible to discover solutions significantly simpler than the best complexifying solutions.

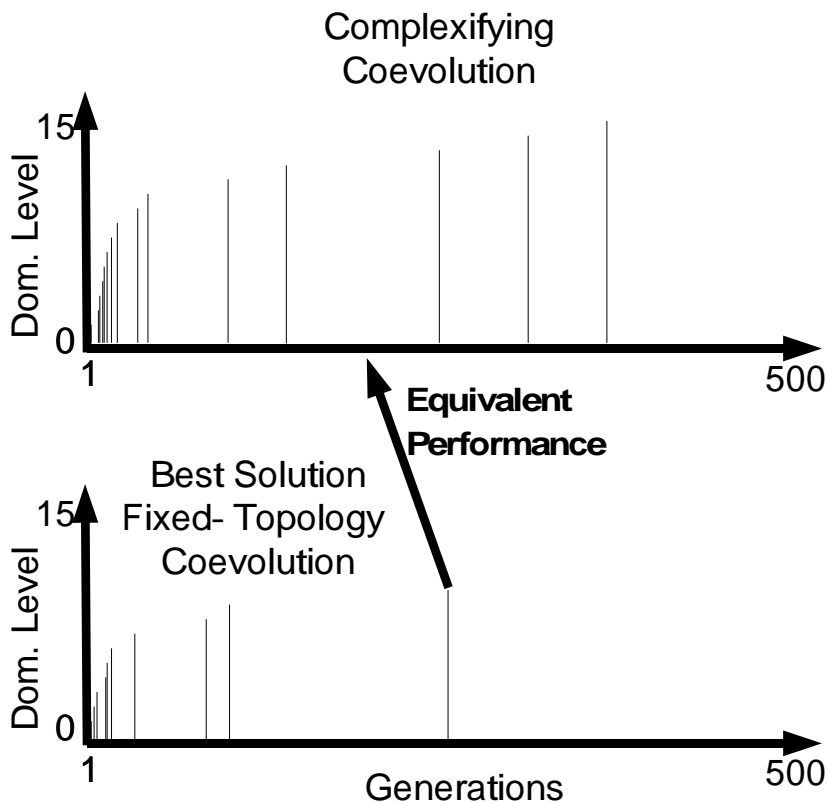

Figure 14: **Comparing Typical Runs of Complexifying Coevolution and Fixed-Topology Coevolution of the Best Complexifying Network.** Dominance levels are charted as in figure 12. The arrow shows that the highest dominant strategy found by evolving the fixed topology of the best complexifying network only performs as well as the dominant strategy that would be found in the 193rd generation of complexifying coevolution (Average 193, sd = 85). Thus, even with an appropriate topology given, fixed-topology evolution takes almost twice as long to achieve the same level of performance.

Thus, we chose to start search with a 12-hidden-unit, 339 connection fully-connected fully-recurrent network. Since 162 connections were added to the best complexifying network during evolution, a corresponding simplifying coevolution could discover solutions with 177 connections, or 25 less than the best complexifying network.

Thus, simplify coevolution was run just as complexifying coevolution, except that the initial topology contained 339 connections instead of 39, and structural mutations removed connections instead of adding nodes and connections. If all connections of a node were removed, the node itself was removed. Historical markings and speciation worked as in complexifying NEAT, except that all markings were assigned in the beginning of evolution. (because structure was only removed and never added). A diversity of species of varying complexity developed as before.

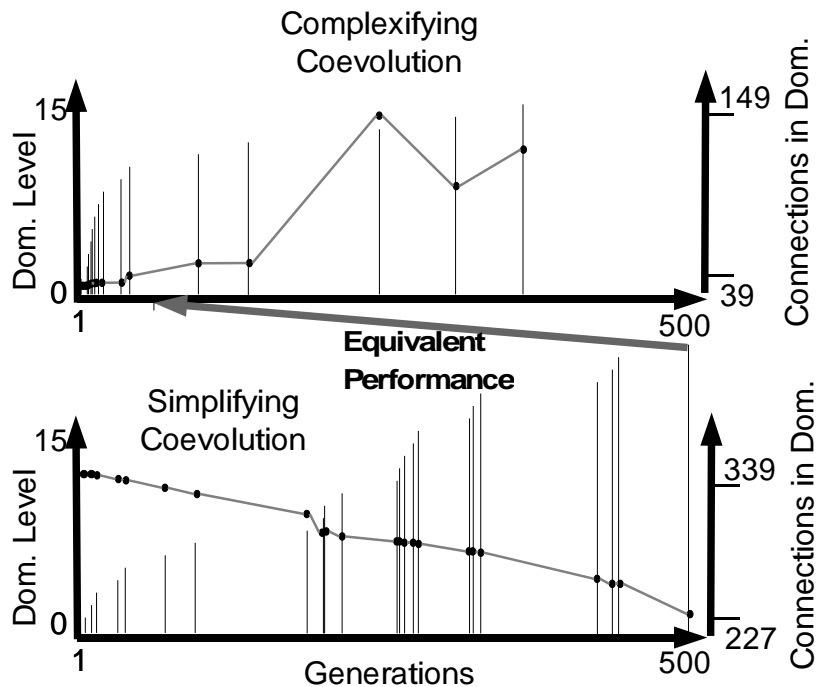

Figure 15: **Comparing Typical Runs of Complexifying Coevolution and Simplifying Coevolution.** Dominance levels are charted as in figure 12. In addition, the line plot shows the complexity of each dominance level in terms of number of connections in the networks with scale indicated in the $y$-axis at right. In this typical simplifying run, the number of connections reduced from 339 to 227 connections. The arrow shows that the highest dominant strategy found in simplifying coevolution only performs as well as the 9th or 10th dominant strategy of complexifying coevolution, which is normally found after 56 generations (sd = 31). In other words, even though simplifying coevolution finds more dominance levels, the search for appropriate structure is less effective than that of complexifying coevolution.

The five runs of simplifying coevolution performed at 64.8%, 60.9%, 56.6%, 36.4%, and 67.9%, with an overall average of 57.3% (sd=19.8%). Again, such performance is significantly below the 91.4% performance of complexifying coevolution ($p < 0.001$). Interestingly, even though it started with 76 more connections than fixed-topology coevolution with ten hidden units, simplifying coevolution still performed significantly better ($p < 0.001$), suggesting that evolving structure through reducing complexity is better than evolving large fixed structures.

Like Figures 12–14, Figure 15 compares typical runs of complexifying and simplifying coevolution. On average, 500 generations of simplification finds solutions equivalent to 56 generations of complexification. Simplifying coevolution also tends to find more dominance levels than any other method tested. It generated an average of 23.2 dominance levels per run, once even finding 30 in one run, whereas e.g. complexifying coevolution on average finds 15.2 levels. In other words, the difference between dominance levels is much smaller in simplifying coevolution than complexifying coevolution. Unlike in other methods, dominant strategies tend to appear in spurts of a few at a time, and usually after complexity has been decreasing for several generations, as also shown in Figure 15. Over a number of generations, evolution removes several connections until a smaller, more easily optimized space is discovered. Then, a quick succession of minute improvements creates several new levels of dominance, after which the space is further refined, and so on. While such a process makes sense, the inferior results of simplifying coevolution suggest that simplifying search is an ineffective way of discovering useful structures compared to complexification.

| Coevolution Type | Ave. Highest Dom. Level | Ave. Highest Generation | Average Performance | Equivalent Generation (out of 500) |
|---|---|---|---|---|
| Complexifying | 15.2 | 353.6 | 91.4% | 343 |
| Fixed-Topology 10 Hidden Node | 12.0 | 172 | 40.4% | 24 |
| Fixed-Topology 5 Hidden Node | 13.0 | 291.4 | 80.3% | 159 |
| Fixed-Topology Best Network | 14.0 | 301.8 | 82.4% | 193 |
| Simplifying | 23.2 | 444.2 | 57.3% | 56 |

Table 1: **Summary of the performance comparison.** The second column shows how many levels of dominance were achieved in each type of coevolution on average. The third specifies the average generation of the highest dominant strategy, indicating how long innovation generally continues. The fourth column gives the average level in the complexifying coevolution dominance hierarchy that the champion could defeat, and the fifth column shows its average generation. The differences in performance ($p < 0.001$) and equivalent generation ($p < 0.001$) between complexifying coevolution and every other method are significant. The main result is that the level of sophistication reached by complexifying coevolution is significantly higher than that reached by fixed-topology or simplifying coevolution.

### 6.3.6 COMPARISON SUMMARY

Table 1 shows how the coevolution methods differ on number of dominance levels, generation of the highest dominance level, overall performance, and equivalent generation. The conclusion is that complexifying coevolution innovates longer and finds a higher level of sophistication than the other methods.

## 7. Discussion and Future Work

What makes complexification such a powerful search method? Whereas in fixed-topology coevolution, as well as in simplifying coevolution, the good structures must be optimized in the high-dimensional space of the solutions themselves, complexifying coevolution only searches high-dimensional structures that are elaborations of known good lower-dimensional structures. Before adding a new dimension, the values of the existing genes have already been optimized over preceding generations. Thus, after a new gene is added, the genome is *already* in a promising part of the new, higher-dimensional space. Thus, the search in the higher-dimensional space is not starting blindly as it would if evolution began searching in that space. It is for this reason that complexification can find high-dimensional solutions that fixed-topology coevolution and simplifying coevolution cannot.

Complexification is particularly well suited for coevolution problems. When a fixed genome is used to represent a strategy, that strategy can be optimized, but it is not possible to add functionality without sacrificing some of the knowledge that is already present. In contrast, if new genetic material can be added, sophisticated elaborations can be layered above existing structure, establishing an evolutionary arms race. This process was evident in the robot duel domain, where successive dominant strategies often built new functionality on top of existing behavior by adding new nodes and connections.

The advantages of complexification do not imply that fixed-sized genomes cannot sometimes evolve increasingly complex *phenotypic behavior*. Depending on the mapping between the genotype and the phenotype, it may be possible for a fixed, finite set of genes to represent solutions (phenotypes) with varying behavioral complexity. For example, such behaviors have been observed in Cellular Automata (CA), a computational structure consisting of a lattice of cells that change their state as a function of their own current state and the state of other cells in their neighborhood. This neighborhood function can be represented in a genome of size $2^{n+1}$ (assuming $n$ neighboring cells with binary state) and evolved to obtain desired target behavior. For example, Mitchell et al. (1996) were able to evolve neighborhood functions to determine whether black or white cells were in the majority in the CA lattice. The evolved CAs displayed complex global behavior patterns that converged on a single classification, depending on which cell type was in the majority. Over the course of evolution, the behavioral complexity of the CA rose even as the genome remained the same size.

In the CA example, the correct neighborhood size was chosen a priori. This choice is difficult to make, and crucial for success. If the desired behavior had not existed within the chosen size, even if the behavior would become gradually more complex, the system would never solve the task. Interestingly, such a dead-end could be avoided if the neighborhood (i.e. the genome) could be expanded during evolution. It is possible that CAs could be more effectively evolved by complexifying (i.e. expanding) the genomes, and speciating to protect innovation, as in NEAT.

Moreover, not only can the chosen neighborhood be too small to represent the solution, but it can also be unnecessarily large. Searching in a space of more dimensions than necessary can impede progress, as discussed above. If the desired function existed in a smaller neighborhood it could have been found with significantly fewer evaluations. Indeed, it is even possible that the most efficient neighborhood is not symmetric, or contains cells that are not directly adjacent to the cell being processed. Moreover, even the most efficient neighborhood may be too large a space in which to begin searching. Starting search in a small space and incrementing into a promising part of higher-dimensional space is more likely to find a solution. For these reasons, complexification can be an advantage, *even if* behavioral complexity can increase to some extent within a fixed space.

The CA example raises the intriguing possibility that *any* structured phenotype can be evolved through complexification from a minimal starting point, historical markings, and the protection of innovation through speciation. In addition to neural networks and CA, electrical circuits (Miller et al., 2000a; Miller, Job, & Vassilev, 2000b), genetic programs (Koza, 1992), robot body morphologies (Lipson & Pollack, 2000), Bayesian networks (Mengshoel, 1999), finite automata (Brave, 1996), and building and vehicle architectures (O'Reilly, 2000) are all structures of varying complexity that can benefit from complexification. By starting search in a minimal space and adding new dimensions incrementally, highly complex phenotypes can be discovered that would be difficult to find if search began in the intractable space of the final solution, or if it was prematurely restricted to too small a space.

The search for optimal structures is a common problem in Artificial Intelligence (AI). For example, Bayesian methods have been applied to learning model structure (Attias, 2000; Ueda & Ghahramani, 2002). In these approaches, the posterior probabilities of different structures are computed, allowing overly complex or simplistic models to be eliminated. Note that these approaches are not aimed at generating increasingly complex functional structures, but rather at providing a model that explains existing data. In other cases, solutions involve growing gradually larger structures, but the goal of the growth is to form gradually better *approximations*. For example, methods like Incremental Grid Growing (Blackmore & Miikkulainen, 1995), and Growing Neural Gas (Fritzke, 1995) add neurons to a network until it approximates the topology of the input space reasonably well. On the other hand, complexifying systems do not have to be non-deterministic (like NEAT), nor do they need to be based on evolutionary algorithms. For example, Harvey (1993) introduced a deterministic algorithm where the chromosome lengths of the entire population increase all at the same time in order to expand the search space; Fahlman & Lebiere (1990) developed a supervised (non-evolutionary) neural network training method called cascade correlation, where new hidden neurons are added to the network in a predetermined manner in order to complexify the function it computes. The conclusion is that complexification is an important general principle in AI.

In the future, complexification may help with the general problem of finding the appropriate level of abstraction for difficult problems. Complexification can start out with a simple, high-level description of the solution, composed of general-purpose elements. If such an abstraction is insufficient, it can be elaborated by breaking down each high-level element into lower level and more specific components. Such a process can continue indefinitely, leading to increasingly complex substructures, and increasingly low-level solutions to subproblems. Although in NEAT the solutions are composed of only connections and nodes, it does provide an early example of how such a process could be implemented.

One of the primary and most elusive goals of AI is to create systems that *scale up*. In a sense, complexification *is* the process of scaling up. It is the general principle of taking a simple idea and elaborating it for broader application. Much of AI is concerned with search, whether over complex multi-dimensional landscapes, or through highly-branching trees of possibilities. However, intelligence is as much about deciding *what space* to search as it is about searching once the proper space has already been identified. Currently, only humans are able to decide the proper level of abstraction for solving many problems, whether it be a simple high-level combination of general-purpose parts, or an extremely complex assembly of low-level components. A program that can decide what level of abstraction is most appropriate for a given domain would be a highly compelling demonstration of Artificial

Intelligence. This is, we believe, where complexification methods can have their largest impact in the future.

## 8. Conclusion

The experiments presented in this paper show that complexification of genomes leads to continual coevolution of increasingly sophisticated strategies. Three trends were found in the experiments: (1) As evolution progresses, complexity of solutions increases, (2) evolution uses complexification to elaborate on existing strategies, and (3) complexifying coevolution is significantly more successful in finding highly sophisticated strategies than non-complexifying coevolution. These results suggest that complexification is a crucial component of a successful search for complex solutions.

## Acknowledgments

This research was supported in part by the National Science Foundation under grant IIS-0083776 and by the Texas Higher Education Coordinating Board under grant ARP-003658-476-2001. Special thanks to an anonymous reviewer for constructive suggestions for non-complexifying comparisons.

## Appendix A. NEAT System Parameters

Each population had 256 NEAT networks, for a total of 512. The coefficients for measuring compatibility were $c_1 = 1.0$, $c_2 = 1.0$, and $c_3 = 2.0$. The initial compatibility distance was $\delta_t = 3.0$. However, because population dynamics can be unpredictable over hundreds of generations, we assigned a target of 10 species. If the number of species grew above 10, $\delta_t$ was increased by 0.3 to reduce the number of species. Conversely, if the number of species fell below 10, $\delta_t$ was decreased by 0.3 to increase the number of species. The normalization factor $N$ used to compute compatibility was fixed at one. In order to prevent stagnation, the lowest performing species over 30 generations old was not allowed to reproduce. The champion of each species with more than five networks was copied into the next generation unchanged. There was an 80% chance of a genome having its connection weights mutated, in which case each weight had a 90% chance of being uniformly perturbed and a 10% chance of being assigned a new random value. (The system is tolerant to frequent mutations because of the protection speciation provides.) There was a 75% chance that an inherited gene was disabled if it was disabled in either parent. In 40% of crossovers, the offspring inherited the *average* of the connection weights of matching genes from both parents, instead of the connection weight of only one parent randomly. In each generation, 25% of offspring resulted from mutation without crossover. The interspecies mating rate was 0.05. The probability of adding a new node was 0.01 and the probability of a new link mutation was 0.1. We used a modified sigmoidal transfer function, $\varphi(x) = \frac{1}{1+e^{-4.9x}}$, at all nodes. These parameter values were found experimentally, and they follow a logical pattern: Links need to be added significantly more often than nodes, and an average weight difference of 0.5 is about as significant as one disjoint or excess gene. Performance is robust to moderate variations in these values. NEAT software is available in the software section at **http://nn.cs.utexas.edu**.

## Appendix B. Robot Duel Domain Coefficients of Motion

The turn angle $\theta$ is determined as $\theta = 0.24|l-r|$, where $l$ is the output of the left turn neuron, and $r$ is the output of the right turn neuron. The robot moves forward a distance of $1.33f$ on the 600 by 600 board, where $f$ is the forward motion output. These coefficients were calibrated through experimentation to achieve accurate and smooth motion with neural outputs between zero and one.